\documentclass{article}

\usepackage[preprint, nonatbib]{nips_2018}

\usepackage[utf8]{inputenc} 
\usepackage[T1]{fontenc}    
\usepackage{hyperref}       
\usepackage{url}            
\usepackage{booktabs}       
\usepackage{amsfonts}       
\usepackage{nicefrac}       
\usepackage{microtype}      
\usepackage{graphicx}
\usepackage{amsmath}
\usepackage{amssymb}

\usepackage{todonotes}

\title{Towards Sparse Hierarchical Graph Classifiers}

\author{
  C\u{a}t\u{a}lina Cangea$^*$$^1$, Petar Veli\v{c}kovi\'{c}$^*$$^1$, Nikola Jovanovi\'{c}$^2$, {\bf Thomas Kipf$^3$ and Pietro Li\`{o}$^1$} \\
  $^1$University of Cambridge \quad $^2$Faculty of Computing, Union University, Belgrade \\ $^3$University of Amsterdam}

\begin{document}

\maketitle

\begin{abstract}
Recent advances in representation learning on graphs, mainly leveraging graph convolutional networks, have brought a substantial improvement on many graph-based benchmark tasks. While novel approaches to learning node embeddings are highly suitable for node classification and link prediction, their application to \emph{graph} classification (predicting a single label for the entire graph) remains mostly rudimentary, typically using a single global pooling step to aggregate node features or a hand-designed, fixed heuristic for hierarchical coarsening of the graph structure. An important step towards ameliorating this is \emph{differentiable graph coarsening}---the ability to reduce the size of the graph in an adaptive, data-dependent manner within a graph neural network pipeline, analogous to image downsampling within CNNs. However, the previous prominent approach to pooling has \emph{quadratic} memory requirements during training and is therefore not scalable to large graphs. Here we combine several recent advances in graph neural network design to demonstrate that competitive hierarchical graph classification results are possible without sacrificing sparsity. Our results are verified on several established graph classification benchmarks, and highlight an important direction for future research in graph-based neural networks.
\end{abstract}

\section{Introduction and Related Work}

Here we{\let\thefootnote\relax\footnote{{$^*$The first two authors contributed equally.}}} study the problem of \emph{graph classification}; the task of learning to categorise graphs into classes. This is a direct generalisation of \emph{image classification} \cite{krizhevsky2012imagenet}, as images may be easily cast as a special case of a ``grid graph'' (with each pixel of an image connected to its eight immediate neighbours). Therefore, it is natural to investigate and generalise CNN elements to graphs \cite{bronstein2017geometric,hamilton2017representation,battaglia2018relational}. 

Generalising the convolutional layer to graphs has been a very active area of research, with several \emph{graph convolutional layers} \cite{bruna2013spectral,defferrard2016convolutional,kipf2016semi,gilmer2017neural,velickovic2018graph} proposed in recent times, significantly advancing the state-of-the-art on many challenging node classification benchmarks (analogues of \emph{image segmentation} in the graph domain), as well as link prediction. 
Conversely, generalising pooling layers has received substantially smaller levels of attention by the community.

The proposed strategies broadly fall into two categories: 1) aggregating node representations in a global pooling step after each \cite{duvenaud2015convolutional} or after the final message passing step \cite{li2015gated,dai2016discriminative,gilmer2017neural}, and 2) aggregating node representations into clusters which coarsen the graph in a hierarchical manner \cite{bruna2013spectral,niepert2016learning,defferrard2016convolutional,monti2017geometric,simonovsky2017dynamic,fey2018splinecnn,mrowca2018flexible,ying2018hierarchical,anonymous2019graph}. Apart from \cite{ying2018hierarchical,anonymous2019graph}, all earlier works in this area assume a fixed, pre-defined cluster assignment, that is obtained by running a clustering algorithm on the graph nodes, e.g.~using the GraClus algorithm \cite{dhillon2007weighted} to obtain structure-dependent cluster assignments or finding clusters via k-means on node features \cite{mrowca2018flexible}. The main insight by recent works \cite{ying2018hierarchical,anonymous2019graph} is that intermediate node representations (e.g.~after applying a graph convolution layer) can be leveraged to obtain both feature- and structure-based cluster assignments that are adaptive to the underlying data and that can be learned in a differentiable manner.

The first end-to-end trainable graph CNN with a learnable pooling operator was recently pioneered, leveraging the DiffPool layer \cite{ying2018hierarchical}. DiffPool computes \emph{soft clustering} assignments of nodes from the original graph to nodes in the pooled graph. Through a combination of restricting the clustering scores to respect the input graph's adjacency information, and a sparsity-inducing entropy regulariser, the clustering learnt by DiffPool eventually converges to an almost-hard clustering with interpretable structure, and leads to state-of-the-art results on several graph classification benchmarks.

The main limitation of DiffPool is computation of the soft clustering assignments---while the assignments eventually converge, during the early phases of training, an entire \emph{assignment matrix} must be stored; relating nodes from the original graph to nodes from the pooled graph in an \emph{all-pairs} fashion. This incurs a \emph{quadratic} $\mathcal{O}(kV^2)$ storage complexity for \emph{any} pooling scheme with a fixed pooling ratio $k$, and is therefore prohibitive for large graphs. 

In this work, we leverage recent advances in graph neural network design \cite{hamilton2017inductive,anonymous2019graph,xu2018representation} to demonstrate that sparsity need not be sacrificed to obtain good performance on end-to-end graph convolutional architectures with pooling. We demonstrate performance that is comparable to variants of DiffPool on four standard graph classification benchmarks, all while using a graph CNN that only requires $\mathcal{O}(V+E)$ storage (comparable to the storage complexity of the input graph).

\section{Model}\label{s2}

We assume a standard graph-based machine learning setup; the input graph is represented as a matrix of \emph{node features}, ${\bf X} \in \mathbb{R}^{N\times F}$, and an \emph{adjacency matrix}, ${\bf A} \in \mathbb{R}^{N\times N}$. Here, $N$ is the number of nodes in the graph, and $F$ the number of features. In the cases where the graph is featureless, one may use the \emph{node degree information} (e.g. one-hot encoding the node degree for all degrees up to a given upper bound) to serve as artificial node features. While the adjacency matrix may consist of \emph{real numbers} (and may even contain \emph{arbitrary edge features}), here we restrict our attention to \emph{undirected} and \emph{unweighted} graphs; i.e. ${\bf A}$ is assumed to be binary and symmetric.

To specify a CNN-inspired neural network for graph classification, we first require a \emph{convolutional} and a \emph{pooling} layer. In addition, we require a \emph{readout} layer (analogous to a \emph{flattening} layer in an image CNN), that converts the learnt representations into a \emph{fixed-size} vector representation, to be used for final prediction (e.g. a simple MLP). These layers are specified in the following paragraphs.

\paragraph{Convolutional layer}
Given that our model will be required to classify \emph{unseen graph structures} at test time, the main requirement of the convolutional layer in our architecture is that it is \emph{inductive}, i.e. that it does not depend on a fixed and known graph structure. The simplest such layer is the \emph{mean-pooling} propagation rule, as similarly used in GCN \cite{kipf2016semi} or Const-GAT \cite{velickovic2018graph}:
\begin{equation}
	\text{MP}({\bf X}, {\bf A}) = \sigma\left({\bf \hat{D}}^{-1}{\bf \hat{A}}{\bf X}{\bf \Theta} + {\bf X}{\bf\Theta'}\right)
\end{equation} 
where ${\bf \hat{A}} = {\bf A} + {\bf I}_N$ is the adjacency matrix with inserted self-loops and $\bf \hat{D}$ is its corresponding degree matrix; i.e. $\hat{D}_{ii} = \sum_j \hat{A}_{ij}$. We have used the rectified linear (ReLU) activation for $\sigma$. ${\bf \Theta}, {\bf \Theta'} \in \mathbb{R}^{F \times F'}$ are learnable linear transformations applied to every node. The transformation through ${\bf\Theta'}$ represents a simple skip-connection \cite{he2016deep}, further encouraging preservation of information about the central node.

\paragraph{Pooling layer}
To make sure that a graph downsampling layer behaves idiomatically with respect to a wide class of graph sizes and structures, we adopt the approach of reducing the graph with a \emph{pooling ratio}, $k \in (0, 1]$. This implies that a graph with $N$ nodes will have $\lceil kN\rceil$ nodes after application of such a pooling layer. 

Unlike DiffPool, which attempts to do this via computing a clustering of the $N$ nodes into $\lceil kN\rceil$ clusters (and therefore incurs a quadratic penalty in storing cluster assignment scores), we leverage the recently proposed Graph U-Net architecture \cite{anonymous2019graph}, which simply \emph{drops} $N - \lceil kN\rceil$ nodes from the original graph. 

The choice of which nodes to drop is done based on a \emph{projection score} against a learnable vector, $\vec{p}$. In order to enable gradients to flow into $\vec{p}$, the projection scores are also used as \emph{gating values}, such that retained nodes receiving lower scores will experience less significant feature retention. Fully written out, the operation of this pooling layer (computing a pooled graph, $\left({\bf X'}, {\bf A'}\right)$, from an input graph, $\left({\bf X}, {\bf A}\right)$), may be expressed as follows:
\begin{equation}
    \vec{y} = \frac{{\bf X}\vec{p}}{\|\vec{p}\|} \qquad \vec{i} = \text{top-$k$}(\vec{y}, k) \qquad {\bf X'} = \left({\bf X} \odot \tanh(\vec{y})\right)_{\vec{i}} \qquad {\bf A'} = {\bf A}_{\vec{i},\vec{i}}
\end{equation}
Here, $\|\cdot\|$ is the $L_2$ norm, $\text{top-$k$}$ selects the top-$k$ indices from a given input vector, $\odot$ is (broadcasted) elementwise multiplication, and $\cdot_{\vec{i}}$ is an indexing operation which takes slices at indices specified by $\vec{i}$. This operation requires only a pointwise projection operation and slicing into the original feature and adjacency matrices, and therefore trivially retains sparsity.

\paragraph{Readout layer}
Lastly, we seek a ``flattening'' operation that will preserve information about the input graph in a fixed-size representation. A natural way to do this in CNNs is \emph{global average pooling}, i.e. the average of all learnt node embeddings in the final layer. We further augment this by performing \emph{global max pooling} as well, which we found strengthened our representations. Lastly, inspired by the JK-net architecture \cite{xu2018representation,xu2018powerful}, we perform this summarisation after \emph{each} conv-pool block of the network, and aggregate all of the summaries together by taking their sum. 

Concretely, to summarise the output graph of the $l$-th conv-pool block, $({\bf X}^{(l)}, {\bf A}^{(l)})$:
\begin{equation}
    \vec{s}^{(l)} = \frac{1}{N^{(l)}}\sum_{i=1}^{N^{(l)}} \vec{x}_i^{(l)} \| \max_{i=1}^{N^{(l)}} \vec{x}_i^{(l)}
\end{equation}
where $N^{(l)}$ is the number of nodes of the graph, $\vec{x}_i^{(l)}$ are the $i$-th node's feature vector, and $\|$ denotes concatenation. Then, the final summary vector (for a graph CNN with $L$ layers) is obtained as the sum of all those summaries (i.e. $\vec{s} = \sum_{l=1}^L \vec{s}^{(l)}$) and submitted to an MLP for obtaining final predictions.

We find that the aggregation across layers is important, not only to preserve information at different scales of processing, but also to handle efficiently retaining information on \emph{smaller} input graphs that may quickly be pooled down to a too small number of nodes.

\begin{figure}[t]
    \centering
    \includegraphics[width=\linewidth]{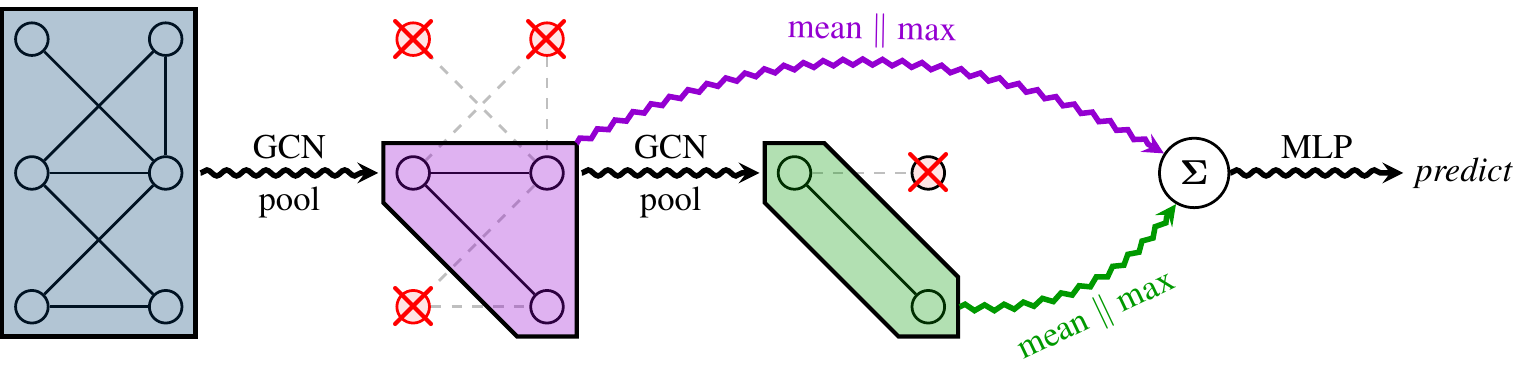}
    \caption{The full pipeline of our model (for $k=0.5$), leveraging several stacks of interleaved convolutional/pooling layers (that, unlike DiffPool, \emph{drop} rather than aggregate nodes), as well as a JK-net-style summary, combining information at different scales.}
    \label{fig:unet}
\end{figure}

The entire pipeline of our model may be visualised in Figure \ref{fig:unet}.

\section{Experiments}

\paragraph{Datasets and evaluation procedure}

To assess how well our \emph{sparse} model can hierarchically compress the representation of a graph while still producing features relevant for classification, we evaluate the graph neural network architecture on several well-known benchmark tasks: biological (\emph{Enzymes}, \emph{Proteins}, \emph{D\&D}) and scientific collaboration (\emph{Collab}) datasets~\cite{KKMMN2016}. We report the performance achieved from carrying out 10-fold cross-validation on each of these, in relation to the results presented by Ying \emph{et al.}~\cite{ying2018hierarchical}.

\paragraph{Model parameters}

Our graph neural network architecture comprises \emph{three} blocks, each of them consisting of a graph convolutional layer with 128 (Enzymes and Collab) or 64 features (D\&D and Proteins), followed by a pooling step (refer to Section~\ref{s2} for details). We ensure that there is enough information after each coarsening stage by preserving 80\% of the existing nodes. A learning rate of 0.005 was used for \emph{Proteins} and 0.0005 for all other datasets. The model was trained using the Adam optimizer~\cite{kingma2014adam} for 100 epochs on \emph{Enzymes}, 40 on \emph{Proteins}, 20 on \emph{D\&D} and 30 on \emph{Collab}.

\paragraph{Results}

\begin{table}
\caption{Classification accuracy percentages. Our model successfully outperforms the sparse aggregation-based GraphSAGE baseline, while being a close competitor to DiffPool variants, across all datasets. This confirms the effectiveness of leveraging learnable pooling while preserving sparsity.}
    \centering
\small
\begin{tabular}{lllll}
 \toprule
  & \multicolumn{4}{c}{\bf{Datasets}} \\
 \cmidrule(r){2-5}
 \bf{Model} & \emph{Enzymes} & \emph{D\&D} & \emph{Collab} & \emph{Proteins} \\
 \midrule
 Graphlet & 41.03 & 74.85 & 64.66 & 72.91 \\
 Shortest-path & 42.32 & 78.86 & 59.10 & 76.43 \\
 1-WL & 53.43 & 74.02 & 78.61 & 73.76 \\
 WL-QA & 60.13 & 79.04 & 80.74 & 75.26 \\
 \midrule
 PatchySAN & – & 76.27 & 72.60 & 75.00 \\
 GraphSAGE & 54.25 & 75.42 & 68.25 & 70.48 \\
 ECC & 53.50 & 74.10 & 67.79 & 72.65 \\
 Set2Set & 60.15 & 78.12 & 71.75 & 74.29 \\
 SortPool & 57.12 & 79.37 & 73.76 & 75.54 \\
 DiffPool-Det & 58.33 & 75.47 & \bf{82.13} & 75.62 \\
 DiffPool-NoLP & 62.67 & 79.98 & 75.63 & 77.42 \\
 DiffPool & \bf{64.23} & \bf{81.15} & 75.50 & \bf{78.10} \\
 \cmidrule(r){1-5}
 Ours & 64.17 & 78.59 & 74.54 & 75.46 \\
 \bottomrule
\label{table:results}
\end{tabular}

\end{table}

\begin{figure}[t]
    \centering
    \includegraphics[width=0.9\linewidth]{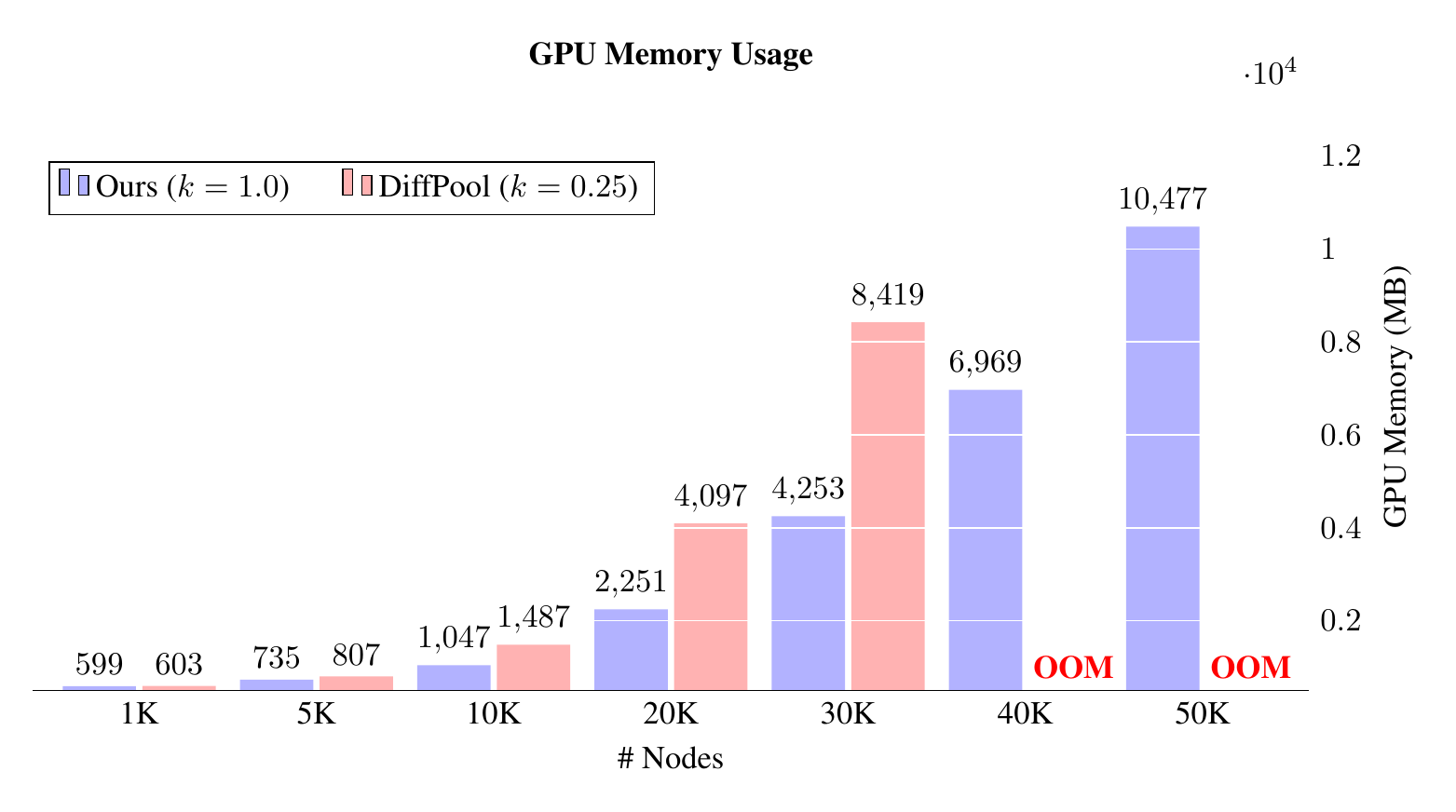}
    \caption{GPU memory usage of our method (with no pooling; $k=1.0$) and DiffPool ($k=0.25$) during training on Erd\H{o}s-R\'{e}nyi graphs \cite{erdos1960evolution} of varying node sizes (and $|E| = 2|V|$). Both methods ran with 128 input and hidden features, and three Conv-Pool layers. ``OOM'' denotes out-of-memory.}
    \label{fig:memo}
\end{figure}

Table~\ref{table:results} illustrates our comparison to the performances reported by Ying \emph{et al.}~\cite{ying2018hierarchical}. In all cases, our algorithm significantly outperforms the GraphSAGE sparse aggregation method~\cite{hamilton2017inductive}, while successfully competing \emph{at most within 1 percentage point of accuracy} with the three variants of DiffPool~\cite{ying2018hierarchical}, the recent singular development in hierarchical graph representation learning. Unlike the latter, our method does not require quadratic memory, paving the way to deploying \emph{scalable} hierarchical graph classification algorithms on larger real-world datasets. 

We also verify this claim empirically---through experiments on random inputs---in Figure \ref{fig:memo}, where we demonstrate that our method compares favourably to DiffPool on larger-scale graphs, even if the pooling layer doesn't drop any nodes (compared to a 0.25 retain rate for the DiffPool).

\section*{Acknowledgements}
We would like to thank the developers of PyTorch \cite{paszke2017automatic}. CC acknowledges funding by DREAM CDT. PV and PL have received funding from the European Union's Horizon 2020 research and innovation programme PROPAG-AGEING under grant agreement No 634821. TK acknowledges funding by SAP SE. We specially thank Jian Tang and Max Welling for the extremely useful discussions.

\bibliographystyle{plain}
\bibliography{nips_2018}

\begin{thebibliography}{10}

\bibitem{anonymous2019graph}
Anonymous.
\newblock Graph u-net.
\newblock In {\em Submitted to the Seventh International Conference on Learning
  Representations (ICLR)}, 2018.
\newblock under review.

\bibitem{battaglia2018relational}
Peter~W Battaglia, Jessica~B Hamrick, Victor Bapst, Alvaro Sanchez-Gonzalez,
  Vinicius Zambaldi, Mateusz Malinowski, Andrea Tacchetti, David Raposo, Adam
  Santoro, Ryan Faulkner, et~al.
\newblock Relational inductive biases, deep learning, and graph networks.
\newblock {\em arXiv preprint arXiv:1806.01261}, 2018.

\bibitem{bronstein2017geometric}
Michael~M Bronstein, Joan Bruna, Yann LeCun, Arthur Szlam, and Pierre
  Vandergheynst.
\newblock Geometric deep learning: going beyond euclidean data.
\newblock {\em IEEE Signal Processing Magazine}, 34(4):18--42, 2017.

\bibitem{bruna2013spectral}
Joan Bruna, Wojciech Zaremba, Arthur Szlam, and Yann LeCun.
\newblock Spectral networks and locally connected networks on graphs.
\newblock {\em arXiv preprint arXiv:1312.6203}, 2013.

\bibitem{dai2016discriminative}
Hanjun Dai, Bo~Dai, and Le~Song.
\newblock Discriminative embeddings of latent variable models for structured
  data.
\newblock In {\em International Conference on Machine Learning}, pages
  2702--2711, 2016.

\bibitem{defferrard2016convolutional}
Micha{\"e}l Defferrard, Xavier Bresson, and Pierre Vandergheynst.
\newblock Convolutional neural networks on graphs with fast localized spectral
  filtering.
\newblock In {\em Advances in Neural Information Processing Systems}, pages
  3844--3852, 2016.

\bibitem{dhillon2007weighted}
Inderjit~S Dhillon, Yuqiang Guan, and Brian Kulis.
\newblock Weighted graph cuts without eigenvectors a multilevel approach.
\newblock {\em IEEE transactions on pattern analysis and machine intelligence},
  29(11), 2007.

\bibitem{duvenaud2015convolutional}
David~K Duvenaud, Dougal Maclaurin, Jorge Iparraguirre, Rafael Bombarell,
  Timothy Hirzel, Al{\'a}n Aspuru-Guzik, and Ryan~P Adams.
\newblock Convolutional networks on graphs for learning molecular fingerprints.
\newblock In {\em Advances in neural information processing systems}, pages
  2224--2232, 2015.

\bibitem{erdos1960evolution}
Paul Erdos and Alfr{\'e}d R{\'e}nyi.
\newblock On the evolution of random graphs.
\newblock {\em Publ. Math. Inst. Hung. Acad. Sci}, 5(1):17--60, 1960.

\bibitem{fey2018splinecnn}
Matthias Fey, Jan~Eric Lenssen, Frank Weichert, and Heinrich M{\"u}ller.
\newblock Splinecnn: Fast geometric deep learning with continuous b-spline
  kernels.
\newblock In {\em Proceedings of the IEEE Conference on Computer Vision and
  Pattern Recognition}, pages 869--877, 2018.

\bibitem{gilmer2017neural}
Justin Gilmer, Samuel~S Schoenholz, Patrick~F Riley, Oriol Vinyals, and
  George~E Dahl.
\newblock Neural message passing for quantum chemistry.
\newblock {\em arXiv preprint arXiv:1704.01212}, 2017.

\bibitem{hamilton2017inductive}
Will Hamilton, Zhitao Ying, and Jure Leskovec.
\newblock Inductive representation learning on large graphs.
\newblock In {\em Advances in Neural Information Processing Systems}, pages
  1024--1034, 2017.

\bibitem{hamilton2017representation}
William~L Hamilton, Rex Ying, and Jure Leskovec.
\newblock Representation learning on graphs: Methods and applications.
\newblock {\em arXiv preprint arXiv:1709.05584}, 2017.

\bibitem{he2016deep}
Kaiming He, Xiangyu Zhang, Shaoqing Ren, and Jian Sun.
\newblock Deep residual learning for image recognition.
\newblock In {\em Proceedings of the IEEE conference on computer vision and
  pattern recognition}, pages 770--778, 2016.

\bibitem{KKMMN2016}
Kristian Kersting, Nils~M. Kriege, Christopher Morris, Petra Mutzel, and Marion
  Neumann.
\newblock Benchmark data sets for graph kernels, 2016.

\bibitem{kingma2014adam}
Diederik Kingma and Jimmy Ba.
\newblock Adam: A method for stochastic optimization.
\newblock {\em arXiv preprint arXiv:1412.6980}, 2014.

\bibitem{kipf2016semi}
Thomas~N Kipf and Max Welling.
\newblock Semi-supervised classification with graph convolutional networks.
\newblock {\em arXiv preprint arXiv:1609.02907}, 2016.

\bibitem{krizhevsky2012imagenet}
Alex Krizhevsky, Ilya Sutskever, and Geoffrey~E Hinton.
\newblock Imagenet classification with deep convolutional neural networks.
\newblock In {\em Advances in neural information processing systems}, pages
  1097--1105, 2012.

\bibitem{li2015gated}
Yujia Li, Daniel Tarlow, Marc Brockschmidt, and Richard Zemel.
\newblock Gated graph sequence neural networks.
\newblock {\em arXiv preprint arXiv:1511.05493}, 2015.

\bibitem{monti2017geometric}
Federico Monti, Davide Boscaini, Jonathan Masci, Emanuele Rodola, Jan Svoboda,
  and Michael~M Bronstein.
\newblock Geometric deep learning on graphs and manifolds using mixture model
  cnns.
\newblock In {\em Proc. CVPR}, volume~1, page~3, 2017.

\bibitem{mrowca2018flexible}
Damian Mrowca, Chengxu Zhuang, Elias Wang, Nick Haber, Li~Fei-Fei, Joshua~B
  Tenenbaum, and Daniel~LK Yamins.
\newblock Flexible neural representation for physics prediction.
\newblock {\em arXiv preprint arXiv:1806.08047}, 2018.

\bibitem{niepert2016learning}
Mathias Niepert, Mohamed Ahmed, and Konstantin Kutzkov.
\newblock Learning convolutional neural networks for graphs.
\newblock In {\em International conference on machine learning}, pages
  2014--2023, 2016.

\bibitem{paszke2017automatic}
Adam Paszke, Sam Gross, Soumith Chintala, Gregory Chanan, Edward Yang, Zachary
  DeVito, Zeming Lin, Alban Desmaison, Luca Antiga, and Adam Lerer.
\newblock Automatic differentiation in pytorch.
\newblock In {\em NIPS-W}, 2017.

\bibitem{simonovsky2017dynamic}
Martin Simonovsky and Nikos Komodakis.
\newblock Dynamic edgeconditioned filters in convolutional neural networks on
  graphs.
\newblock In {\em Proc. CVPR}, 2017.

\bibitem{velickovic2018graph}
Petar Veli{\v{c}}kovi{\'{c}}, Guillem Cucurull, Arantxa Casanova, Adriana
  Romero, Pietro Li{\`{o}}, and Yoshua Bengio.
\newblock {Graph Attention Networks}.
\newblock {\em International Conference on Learning Representations}, 2018.

\bibitem{xu2018powerful}
Keyulu Xu, Weihua Hu, Jure Leskovec, and Stefanie Jegelka.
\newblock How powerful are graph neural networks?
\newblock {\em arXiv preprint arXiv:1810.00826}, 2018.

\bibitem{xu2018representation}
Keyulu Xu, Chengtao Li, Yonglong Tian, Tomohiro Sonobe, Ken-ichi Kawarabayashi,
  and Stefanie Jegelka.
\newblock Representation learning on graphs with jumping knowledge networks.
\newblock {\em arXiv preprint arXiv:1806.03536}, 2018.

\bibitem{ying2018hierarchical}
Rex Ying, Jiaxuan You, Christopher Morris, Xiang Ren, William~L Hamilton, Jure
  Leskovec, Joseph~M Antognini, Jascha Sohl-Dickstein, Nima Roohi, Ramneet
  Kaur, et~al.
\newblock Hierarchical graph representation learning with differentiable
  pooling.
\newblock {\em CoRR}, 2018.

\end{thebibliography}

\appendix

\section{Qualitative analysis}

\begin{figure}[h]
    \centering
    \includegraphics[width=0.6\linewidth]{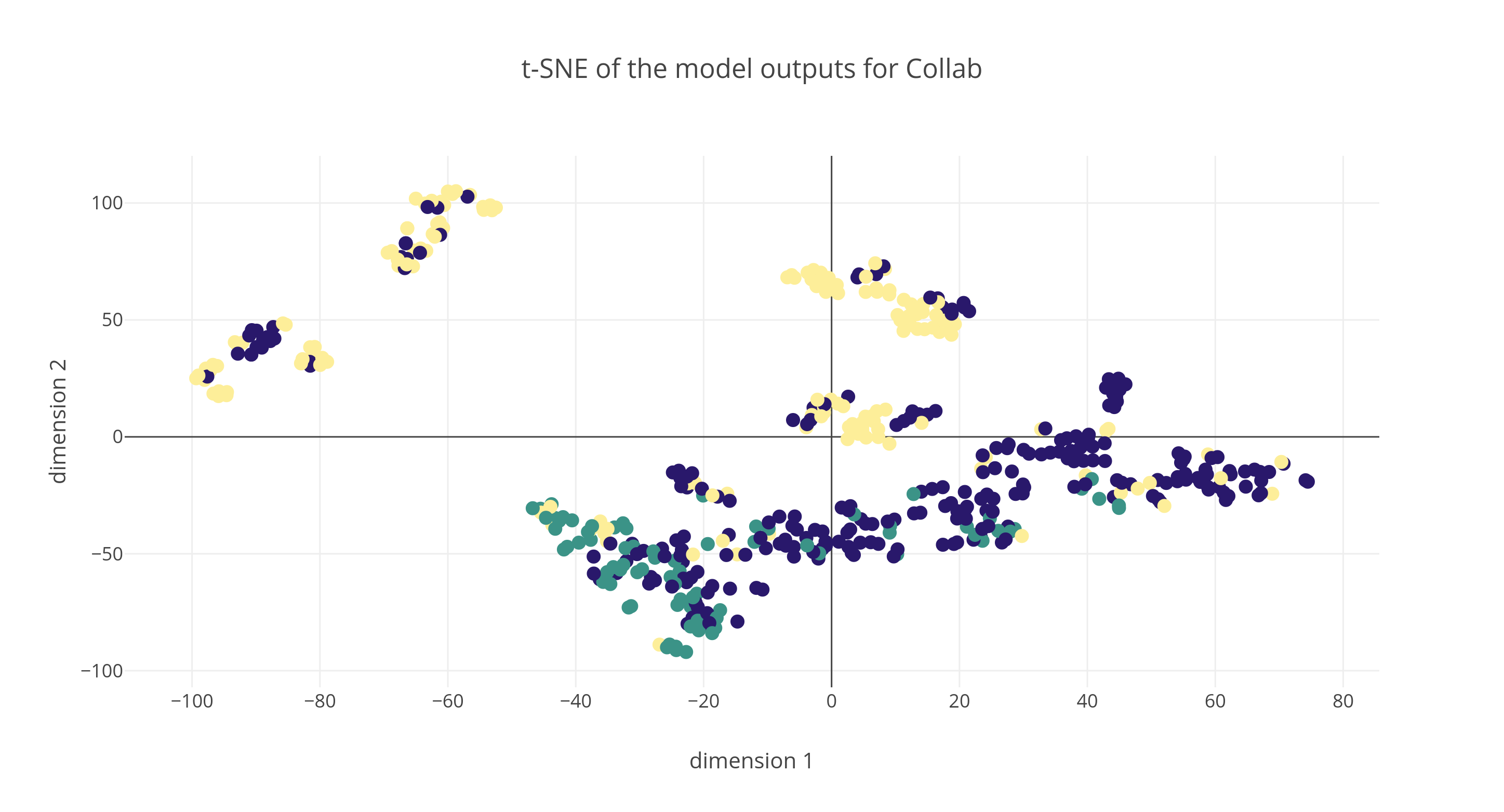}
    \caption{t-SNE plot illustrating the classification capabilities of our model. The points represent summaries of 499 \emph{Collab} test graphs; each of the three classes corresponds to a different color.}
    \label{fig:tsne}
\end{figure}

We qualitatively investigate the distribution of graph summaries, using a pre-trained model on a fold of the \emph{Collab} dataset to produce 499 outputs across all 3 classes. Figure~\ref{fig:tsne} shows that an evident clustering can be achieved, once the graph has been processed by the sequence of convolution and pooling layers leveraged by our architecture.

\end{document}